\documentclass[11pt,a4paper]{article}
\usepackage[hyperref]{emnlp2018}
\usepackage{times}
\usepackage{latexsym}

\usepackage{url}

\usepackage{CJKutf8}
\usepackage{amsmath}
\usepackage{graphicx}
\usepackage{amsfonts}
\usepackage{bm}
\usepackage{multirow}
\aclfinalcopy % Uncomment this line for the final submission
\title{Exploration on Generating Traditional Chinese Medicine Prescriptions from Symptoms with an End-to-End Approach}

\author{Wei Li, Xu Sun\\
MOE Key Laboratory of Computational Linguistics, \\School of Electronics Engineering and Computer \\Science, Peking University\\ {\tt liweitj47,xusun@pku.edu.cn}\\\And
Zheng Yang\\
Beijing University \\of Chinese Medicine\\ {\tt yangzheng@bucm.edu.cn}}

\date{}

\begin{document}
\maketitle
\begin{abstract}
Traditional Chinese Medicine (TCM) is an influential form of medical treatment in China and surrounding areas. In this paper, we propose a TCM prescription generation task that aims to automatically generate a herbal medicine prescription based on textual symptom descriptions. Sequence-to-sequence (seq2seq) model has been successful in dealing with sequence generation tasks. We explore a potential end-to-end solution to the TCM prescription generation task using seq2seq models. However, experiments show that directly applying seq2seq model leads to unfruitful results due to the repetition problem. To solve the problem, we propose a novel decoder with coverage mechanism and a novel soft loss function. The experimental results demonstrate the effectiveness of the proposed approach. Judged by professors who excel in TCM, the generated prescriptions are rated 7.3 out of 10. It shows that the model can indeed help with the prescribing procedure in real life.
\end{abstract}

\setlength{\abovedisplayskip}{3pt}
\setlength{\belowdisplayskip}{3pt}
\setlength{\abovedisplayshortskip}{3pt}
\setlength{\belowdisplayshortskip}{3pt}

\section{Introduction}
\label{sec:Introduction}
Traditional Chinese Medicine (TCM) is one of the most important forms of medical treatment in China and the surrounding areas. TCM has accumulated large quantities of documentation and therapy records in the long history of development. Prescriptions consisting of herbal medication are the most important form of TCM treatment. TCM practitioners prescribe according to a patient's symptoms that are observed and analyzed by the practitioners themselves instead of using medical equipment, e.g., the CT. The patient takes the decoction made out of the herbal medication in the prescription. A complete prescription includes the composition of herbs, the proportion of herbs, the preparation method and the doses of the decoction. In this work, we focus on the composition part of the prescription, which is the most essential part of the prescription.

During the long history of TCM, there has been a number of therapy records or treatment guidelines in the TCM classics composed by outstanding TCM researchers and practitioners. In real life, TCM practitioners often take these classical records for reference when prescribing for the patient, which inspires us to design a model that can automatically generate prescriptions by learning from these classics. It also needs to be noted that due to the issues in actual practice, the objective of this work is to generate candidate prescriptions to facilitate the prescribing procedure instead of substituting the human practitioners completely. % The objective of this work for now is to help the practitioners generate prescriptions instead of replacing the human practitioners.

An example of TCM prescription is shown in Table \ref{tab:TCM prescription example}. The herbs in the prescription are organized in a weak order. By ``weak order'', we mean that the effect of the herbs are not influenced by the order. However, the order of the herbs reflects the way of thinking when constructing the prescription. Therefore, the herbs are connected to each other, and the most important ones are usually listed first. %Another thing that should be noted is that there is no duplicate herb in a prescription, which is different from natural language. 
\begin{table}[ht]
\begin{center}
\small
\begin{tabular}{|c|p{5cm}|}
\hline
Name & \begin{CJK*}{UTF8}{gbsn}麻黄汤\end{CJK*} (Mahuang decoction) \\ \hline
Symptoms & \begin{CJK*}{UTF8}{gbsn}外感风寒表实证。恶寒发热，头身疼痛，无汗自喘，舌苔薄白，脉浮紧。\end{CJK*} \\ \hline
Translation & Affection of exogenous wind-cold; aversion to cold, fever; headache and body pain; adiapneustia and pant; thin and white tongue coating, floating and tense pulse \\ \hline
Prescription & \begin{CJK*}{UTF8}{gbsn}麻黄、桂枝、杏仁、甘草\end{CJK*}  \\ \hline
Translation & Mahuang (ephedra), Guizhi (cassia twig), Xingren (almond), Gancao (glycyrrhiza) \\ \hline
\end{tabular}
\caption{An example of a TCM symptom-prescription pair. As we are mainly concerned with the composition of the prescription, we only provide the herbs in the prescription. \label{tab:TCM prescription example}}
\end{center}
\end{table}
Due to the lack of digitalization and formalization, TCM has not attracted sufficient attention in the artificial intelligence community. To facilitate the studies on automatic TCM prescription generation, we collect and clean a large number of prescriptions as well as their corresponding symptom descriptions from the Internet. 

Inspired by the great success of natural language generation tasks like neural machine translation (NMT) \cite{bahdanau2014neural,cho2014learning,sutskever2014sequence}, abstractive summarization \cite{see2017get}, generative question answering \cite{yin2015neural}, and neural dialogue response generation \cite{li2017adversarial,li2016deep}, we propose to adopt the end-to-end paradigm, mainly the sequence to sequence model, to tackle the task of generating TCM prescriptions based on textual symptom descriptions. 

The sequence to sequence model (seq2seq) consists of an encoder that encodes the input sequence and a decoder that generates the output sequence. The success in the language generation tasks indicates that the seq2seq model can learn the semantic relation between the output sequence and the input sequence quite well. It is also a desirable characteristic for generating prescriptions according to the textual symptom description.

The prescription generation task is similar to the generative question answering (QA). In such task settings, the encoder part of the model takes in the question, and encodes the sequence of tokens into a set of hidden states, which embody the information of the question. The decoder part then iteratively generates tokens based on the information encoded in the hidden states of the encoder. The model would learn how to generate response after training on the corresponding question-answer pairs.

In the TCM prescription generation task, the textual symptom descriptions can be seen as the question and the aim of the task is to produce a set of TCM herbs that form a prescription as the answer to the question. However, the set of herbs is different from the textual answers to a question in the QA task. A difference that is most evident is that there will not be any duplication of herbs in the prescription. 
However, the basic seq2seq model sometimes produces the same herb tokens repeatedly when applied to the TCM prescription generation task. This phenomenon can hurt the performance of recall rate even after applying a post-process to eliminate repetitions. Because in a limited length of the prescription %(including the situation where $<EOS>$ is met)
, the model would produce the same token over and over again, rather than real and novel ones. Furthermore, the basic seq2seq assumes a strict order between generated tokens, but in reality, we should not severely punish the model when it predicts the correct tokens in the wrong order. %On the symptom side, although the symptoms are displayed in the form of natural language, they are strongly connected to each other, which means the symptoms may be talking about the same disease from different aspects.

In this paper, we explore to automatically generate TCM prescriptions based on textual symptoms. We propose a soft seq2seq model with coverage mechanism and a novel soft loss function. The coverage mechanism is designed to make the model aware of the herbs that have already been generated while the soft loss function is to relieve the side effect of strict order assumption. In the experiment results, our proposed model beats all the baselines in professional evaluations, and we observe a large increase in both the recall rate and the F1 score compared with the basic seq2seq model.

The main contributions of this paper lie in the following three folds:
\begin{itemize}
\item We propose a TCM prescription generation task and collect a large quantity of TCM prescription data including symptom descriptions. It is the first time that this task has been considered to our knowledge. 
\item We propose to apply an end-to-end method to deal with the TCM prescription generation problem. In the experiments, we observe that directly applying seq2seq model would result in low recall rate because of the repetition problem.
\item We propose to enhance the basic seq2seq model with cover mechanism and soft loss function to guide the model  to generate more fruitful results. In our experiments, the professional human evaluation score reaches 7.3 (out of 10), which shows that our model can indeed help the TCM practitioners to prescribe in real life. Our final model also increases the F1 score and the recall rate in automatic evaluation by a substantial margin compared with the basic seq2seq model.  
\end{itemize}

\section{Related Work}
%\subsection{Computational TCM Methods}
There has not been much work concerning computational TCM. \newcite{zhou2010development} attempted to build a TCM clinical data warehouse so that the TCM knowledge can be analyzed and used. This is a typical way of collecting data, since the number of prescriptions given by the practitioners in the clinics is very large. However, in reality, most of the TCM doctors do not refer to the constructed digital systems, because the quality of the input data tends to be poor. Therefore, we choose prescriptions in the classics (books or documentation) of TCM. Although the available data can be fewer than the clinical data, it guarantees the quality of the prescriptions. 

\newcite{wang2004self} attempted to construct a self-learning expert system with several simple classifiers to facilitate the TCM diagnosis procedure, \newcite{Wang2013TCM} proposed to use shallow neural networks and CRF based multi-labeling learning methods to model TCM inquiry process, but they only considered the disease of chronic gastritis and its taxonomy is very simple. These methods either utilize traditional data mining methods or are highly involved with expert crafted systems. \newcite{Zhang2011Topic,Zhu2017TCM} proposed to use LDA to model the herbs. \newcite{li2017distributed} proposed to learn the distributed embedding for TCM herbs with recurrent neural networks.

\section{Methodology}
\label{sec:Model}
\begin{figure*}[t]
\centering
\includegraphics[scale=0.6]{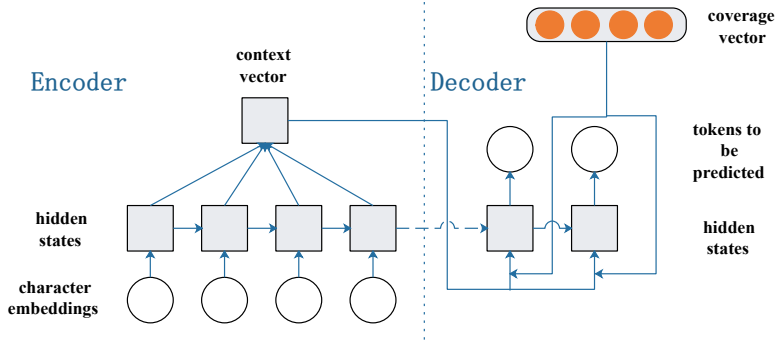}
\caption{An illustration of our model. The model is built on the basis of seq2seq model with attention mechanism. We use a coverage mechanism to reduce repetition problem. The coverage mechanism is realized by adding a coverage vector to the decoder.\label{Model Figure}}
\end{figure*}

Neural sequence to sequence model has proven to be very effective in a wide range of natural language generation tasks, including neural machine translation and abstractive text summarization. In this section, we first describe the definition of the TCM prescription generation task. Then, we introduce how to apply seq2seq model in the prescription composition task. Next, we show how to guide the model to generate more fruitful herbs in the setting of this task by introducing coverage mechanism. Finally, we introduce our novel soft loss function that relieves the strict assumption of order between tokens. An overview of the our final model is shown in Figure \ref{Model Figure}.

\subsection{Task Definition}
Given a TCM herbal treatment dataset that consists of $N$ data samples, the $i$-th data sample ($x^{(i)}, p^{(i)}$) contains one piece of source text $x^{(i)}$  that describes the symptoms, and $M_{i}$ TCM herbs $(p_{1}^{i},p_{2}^{i}, ..., p_{M_{i}}^{i})$ that make up the herb prescription $p^{(i)}$. 

We view the symptoms as a sequence of characters $x^{(i)} = (x^{(i)}_{1}, x^{(i)}_{2}, ..., x^{(i)}_{T})$. We do not segment the characters into words because they are mostly in traditional Chinese that uses characters as basic semantic units. The herbs $p_{1}^{i},p_{2}^{i}, ..., p_{M_{i}}^{i}$ are all different from each other. 

\subsection{Basic Encoder-Decoder Model}
Sequence-to-sequence model was first proposed to solve the machine translation problem. The model consists of two parts, an encoder and a decoder. The encoder is bound to take in the source sequence and compress the sequence into a series of hidden states. The decoder is used to generate a sequence of target tokens based on the information embodied in the hidden states given by the encoder. Typically, both the encoder and the decoder are implemented with recurrent neural networks (RNN). 

In our TCM prescription generation task, the encoder RNN converts the variable-length symptoms in character sequence $x = (x_{1},x_{2},...,x_{T})$ into a set of hidden representations $h = (h_{1},h_{2},...,h_{T})$, by iterating the following equations along time $t$:
\begin{equation}
h_{t} = f(x_{t},h_{t-1})
\end{equation}
where $f$ is a RNN family function. In our implementation, we choose gated recurrent unit (GRU \cite{cho2014learning}) as $f$, as the gating mechanism is expected to model long distance dependency better. Furthermore, we choose the bidirectional version of recurrent neural networks as the encoder to solve the problem that the later words get more emphasis in the unidirectional version. We concatenate both the $h_{t}$ in the forward and backward pass and get $\widehat{h_{t}}$ as the final representation of the hidden state at time step $t$.

We get the context vector $c$ representing the whole source $x$ at the $t$-th time through a non-linear function $q$, normally known as the attention mechanism:
\begin{eqnarray}
c_{t} = \sum_{j=1}^{T}\alpha_{tj}h_{j} \\
\alpha_{tj} = \frac{\text{exp}\left( a\left(s_{t-1},h_{j}\right)\right)}{\sum_{k=1}^{T}\text{exp}\left( a\left(s_{t-1},h_{k}\right)\right)}
\end{eqnarray}
The context vector $c_{t}$ is calculated as a weighted sum of hidden representation produced by the encoder $\textbf{h} = (h_{1},...,h_{T})$. $a(s_{t-1},h_{j})$ is a soft alignment function that measures the relevance between $s_{t-1}$ and $h_{j}$. It computes how much $h_j$ is needed for the $t$-th output word based on the previous hidden state of the decoder $s_{t-1}$. %To put it more clearly, function $a$ is expected to measure to which degree the decoder needs to extract information from $j$-th time step in the hidden state of the encoder. 

The decoder is another RNN. It generates a variable-length sequence $y = (y_{1},y_{2}, ..., y_{T'})$ token by token (herb), through a conditional language model:
\begin{eqnarray}
s_{t} = f(s_{t-1},c_{t},Ey_{t-1}) \label{equ:gru_equ} \\
p(y_{t}|y_{1,...,t},x) = g(s_{t})
\end{eqnarray}
where $s_{t}$ is the hidden state of the decoder RNN at time step $t$. $f$ is also a gated recurrent unit. The non-linear function $g$ is a $softmax$ layer, which outputs the probabilities of all the herbs in the herb vocabulary. $E \in (V\times d)$ is the embedding matrix of the target tokens, $V$ is the number of herb vocabulary, $d$ is the embedding dimension. $y_{t-1}$ is the last predicted token.

In the decoder, the context vector $c_{t}$ is calculated based on the hidden state $s_{t-1}$ of the decoder at time step $t-1$ and all the hidden states in the encoder. The procedure is known as the attention mechanism. The attention mechanism is expected to supplement the information from the source sequence that is more connected to the current hidden state of the decoder instead of only depending on a fixed vector produced by the encoder.

The encoder and decoder networks are trained jointly to maximize the conditional probability of the target sequence. A soft version of cross entropy loss is applied to maximize the conditional probability, which we will describe in detail.

\subsection{Coverage Mechanism}
Different from natural language generation tasks, there is no duplicate herb in the TCM prescription generation task. When directly applying seq2seq model in this task, the decoder tends to generate some frequently observed herbs over and over again. Although we can prune the repeated herbs through post processing by eliminating the repeated ones, it still hurts the recall performance as the maximum length of a prescription is limited. This situation is still true when we use a $<EOS>$ label to indicate where the generation should stop.

To encourage the decoder to generate more diverse and reasonable herb tokens, we propose to apply coverage mechanism to make the model aware of the already generated herbs. Coverage mechanism \cite{tu2016modeling,Tu2016,Mi2016} was first proposed to help the decoder focus on the part that has not been paid much attention by feeding a fertility vector to the attention calculation, indicating how much information of the input is used. 

In our model, we do not use the fertility vector to tune the attention weights. The reason is that the symptoms are related to others and altogether describe the whole disease, which is explained in Section \ref{sec:Introduction}. Still, inspired by its motivation, we adapt the coverage mechanism to the decoder where a coverage vector is fed to the GRU cell together with the context vector. Equation \ref{equ:gru_equ} is then replaced by the following ones.
\begin{eqnarray}
a_{t} = \tanh (WD_{t}+b) \\
s_{t} = f(s_{t-1}, c_{t}, Ey_{t-1}, a_{t})
\end{eqnarray}
where $a_{t}$ is the coverage vector at the $t$-th time step in decoding. $D_{t}$ is the one-hot representation of the generated tokens until the $t$-th time step. $W\in \mathbb{R}^{V\times H}$ is a learnable parameter matrix, where $V$ is the size of the herb vocabulary and $H$ is the size of the hidden state. By feeding the coverage vector, which is also a sketch of the generated herbs, to the GRU as part of the input, our model can softly switch more probability to the herbs that have not been predicted. This way, the model is encouraged to produce novel herbs rather than repeatedly predicting the frequently observed ones, thus increasing the recall rate.

\subsection{Soft Loss Function}
We argue that even though the order of the herbs matters when generating the prescription \citep{vinyals2015order,nam2017maximizing}, we should not strictly restrict the order. However, the traditional cross entropy loss function applied to the basic seq2seq model puts a strict assumption on the order of the labels. To deal with the task of predicting weakly ordered labels (or even unordered labels), we propose a soft loss function instead of the original hard cross entropy loss function:
\begin{equation}
loss = -\sum_{t}\ q'_{t}\ log(p_t)
\end{equation}
Instead of using the original hard one-hot target probability $q_t$, we use a soft target probability distribution $q'_{t}$, which is calculated according to $q_t$ and the target sequence $\bm{q}$ of this sample. Let $\bm{q_v}$ denote the bag of words representation of $\bm{q}$, where only slots of the target herbs in $\bm{q}$ are filled with $1s$. We use a function $\xi$ to project the original target label probability $q_t$ into a new probability distribution $q'_t$.
\begin{equation}
q'_t = \xi(q_t, \bm{q_v})
\end{equation}
This function $\xi$ is designed so as to decrease the harsh punishment when the model predicts the labels in the wrong order. In this paper, we apply a simple yet effective projection function as Equation \ref{equ:soft_loss}. This is an example implementation, and one can design more sophisticated projection functions if needed.
\begin{equation}
\xi(y_t,\bm{s}) = ((\bm{q_v}/M) + y_t) / 2 \label{equ:soft_loss}
\end{equation}
where $M$ is the length of $q$. This function means that at the $t$-th time of decoding, for each target herb token $p_i$, we first split a probability density of $1.0$ equally across all the $l$ herbs into $1/M$. Then, we take the average of this probability distribution and the original probability $q_t$ to be the final probability distribution at time $t$.

%Suppose the overall target sequence is $(s_i,s_j,s_k)$, and there are $V$ labels in the label space in total. At the first time step, the original target probability distribution would be $(0,0,\cdots,\bm{1},\cdots,0)$, where the only ``1'' appears at the $s_i$-th slot in the one-hot representation. In our soft loss function, we convert the one-hot distribution into a soft probability distribution, $\xi(0,0,\cdots,\bm{1},\cdots,0) = (0,0,\cdots,\bm{0.666},0,\cdots,\bm{0.167},0,\cdots,\bm{0.167},0,\cdots,0)$, where the $s_i$-th slot is $0.666$, and the $s_j$,$s_k$-th slots are equally set to $0.167$. The probability of the supposed herb is still in the dominating position in the new distribution ($0.666 > 0.167$), but the herbs that appear in other slots are also allocated with some probability ($0.167$) instead of $0$. 

\section{Experiment}
\subsection{Dataset Construction}
\label{sec:data_construction}
%When constructing our TCM herbal therapy dataset, we first considered the TCM medical records \begin{CJK*}{UTF8}{gbsn}(中医医案)\end{CJK*} in the history, which contain a lot of good reference medical cases. The medical records are widely referenced by the doctors in the treatment. However, they have not been well digitalized, which makes it hard to extract the prescriptions out of the descriptive natural language text from the records. Another way to get large scale prescriptions is to collect from TCM clinics. The problem is that this kind of valuable data is not publicly available and the quality of such data is not good enough. Therefore, we turned to classic resources on the Internet finally. 

We crawl the data from \begin{CJK*}{UTF8}{gbsn}TCM Prescription Knowledge Base\  (中医方剂知识库)\end{CJK*} \footnote{\url{http://www.hhjfsl.com/fang/}}. This knowledge base includes comprehensive TCM documentation in the history. The database includes 710 TCM historic books or documents as well as some modern ones, consisting of 85,166 prescriptions in total. Each item in the database provides the name, the origin, the composition, the effect, the contraindications, and the preparation method. We clean and formalize the database and get 82,044 usable symptom-prescription pairs  

In the process of formalization, we temporarily omit the dose information and the preparation method description, as we are mainly concerned with the composition. Because the names of the herbs have evolved a lot, we conclude heuristic rules as well as specific projection rules to project some rarely seen herbs to their similar forms that are normally referred to. There are also prescriptions that refer to the name of other prescriptions. We simply substitute these names with their constituents.

To make the experiment result more robust, we conduct our experiments on two separate test datasets. The first one is a subset of the data described above. We randomly split the whole data into three parts, the training data (90\%), the development data (5\%) and the test data (5\%). The second one is a set of symptom-prescription pairs we manually extracted from the modern \textbf{text book} of the course \textbf{Formulaology of TCM} (\begin{CJK*}{UTF8}{gbsn}中医方剂学\end{CJK*}) that is popularly adopted by many TCM colleges in China.

There are more cases in the first sampled test dataset (4,102 examples), but it suffers from lower quality, as this dataset was parsed with simple rules, which may not cover all exceptions. The second test dataset has been proofread and all of the prescriptions are the most classical and influential ones in the history. So the quality is much better than the first one. However, the number of the cases is limited. There are 141 symptom-prescription pairs in the second dataset. Thus we use two test sets to do evaluation to take the advantages of both data magnitude and quality.

\subsection{Experiment Settings}
\begin{table}[htb]
\centering
\begin{tabular}{|c|c|c|c|}
\hline
Data & Average & Max & Under 20 \\ \hline
Crawled Data & 7.2 & 108 & 97.99\% \\
Textbook Data & 6.7 & 16 & 100\% \\ \hline
\end{tabular}
\caption{The statistic of the \textbf{length} of prescriptions. Crawled data means the overall data crawled from the Internet, including the training set data, the development set data and test set 1. Textbook data is the same to test set 2. Under 20 means the percentage of data that are shorter or equal than length 20. }\label{tab:data length distribution}
\end{table}
In our experiments, we implement our models with the PyTorch toolkit \footnote{\url{www.pytorch.org}}. We set the embedding size of both Chinese characters in the symptoms and the herb tokens to 100. We set the hidden state size to 300, and the batch size to 20. We set the maximum length of the herb sequence to 20 because the length of nearly all the prescriptions are within this range (see Table \ref{tab:data length distribution} for the statistics of the length of prescriptions). % 16, 6.7; 99, 7.3, 97.99%
Unless specifically stated, we use bidirectional gated recurrent neural networks (BiGRNN) to encode the symptoms. Adam \cite{Kingma2015}, and use the model parameters that generate the best F1 score on the development set in testing

\subsection{Proposed Baseline}
\label{sec:baseline}
In this sub-section, we present the \textbf{Multi-label} baseline we apply. In this model, we use a BiGRNN as the encoder, which encodes symptoms in the same way as it is described in Section \ref{sec:Model}. Because the position of the herbs does not matter in the results, for the generation part, we implement a multi-label classification method to predict the herbs. We use the multi-label max-margin loss (MultiLabelMarginLoss in pytorch) as the optimization objective, because this loss function is more insensitive to the threshold, thus making the model more robust. We set the threshold to be 0.5, that is, if the probability given by the model is above 0.5 and within the top $k$ range (we set k to 20 in our experiment, same to seq2seq model), we take the tokens as answers. The way to calculate probability is shown below. 
\begin{equation}
p(i) = \sigma(W_{o}h_{T})
\end{equation}
where $\sigma$ indicates the non-linear function $sigmoid$, $W_{o} \in \mathbb{R}^{H \times V}$, $H$ is the size of the hidden state produced by the encoder and $V$ is the size of the herb vocabulary. $h_{T}$ is the last hidden state produced by the encoder.

During evaluation, we choose the herbs satisfying two conditions:
\begin{enumerate}
\item The predicted probability of the herb is within top $k$ among all the herbs, where $k$ is a hyper-parameter. We set $k$ to be the same as the maximum length of seq2seq based models (20).
\item The predicted probability is above a threshold 0.5 (related to the max-margin).
\end{enumerate}

\subsection{Human Evaluation}
Since medical treatment is a very complex task, we invite two professors from Beijing University of Chinese Medicine, which is one of the best Traditional Chinese Medicine academies in China. Both of the professors enjoy over five years of practicing traditional Chinese medical treatment. The evaluators are asked to evaluate the prescriptions with scores between 0 and 10. Both the textual symptoms and the standard reference are given, which is similar to the form of evaluation in a normal TCM examination.  Different from the automatic evaluation method, the human evaluators focus on the potential curative effect of the candidate answers, rather than merely the literal similarity. We believe this way of evaluation is much more reasonable and close to reality. 

Because the evaluation procedure is very time consuming (each item requires more than 1 minute), we only ask the evaluators to judge the results from test set 2.

\begin{table}[tb]
\centering
\begin{tabular}{|l|c|c|c|}
\hline
Model & E 1 & E 2 & Average \\ \hline
Multi-Label &4.5 & 4.1 & 4.3 \\
Basic seq2seq & 6.8 & 6.6 & 6.7  \\
Proposal & 7.4 & 7.1 & 7.3  \\ \hline
\end{tabular}
\caption{Professional evaluation on the test set 2. The score range is 0$\sim$10. The Pearson's correlation coefficient between the two evaluators is 0.72 and the Spearman's correlation coefficient is 0.72. Both p-values are less than 0.01, indicating strong agreement.}\label{tab: Professional Evaluation}
\end{table}

As shown in Table \ref{tab: Professional Evaluation}, both of the basic seq2seq model and our proposed modification are much better than the multi-label baseline. Our proposed model gets a high score of 7.3, which  can be of real help to TCM practitioners when prescribing in the real life treatment. 

\subsection{Automatic Evaluation Results}
We use micro Precision, Recall, and F1 score as the automatic metrics to evaluate the results, because the internal order between the herbs does not matter when we do not consider the prescribing process.
\begin{table*}[tb]
\centering
\begin{tabular}{|l|c|c|c|c|c|c|}
\hline
\multirow{2}{*}{Model} & \multicolumn{3}{c|}{Test set 1} & \multicolumn{3}{c|}{Test set 2} \\ \cline{2-7}
& P & R & F & P & R & F \\ \hline
Multi-label & 10.83 & 29.72 & 15.87 & 13.51 & 40.49 & 20.26 \\  
Basic seq2seq & 26.03 & 13.52 & 17.80 & 30.97 & 23.70 & 26.85 \\ \hline
Proposal & 29.57 & 17.30 & 21.83 & 38.22 & 30.18 & 33.73 \\  \hline
\end{tabular}
\caption{Automatic evaluation results of different models on the two test datasets. Multi-label is introduced in Section \ref{sec:baseline}. Test set 1 is the subset of the large dataset collected from the Internet, which is homogeneous to the training set. Test set 2 is the test set extracted from the prescription text book.}\label{tab:micro Experiment results}
\end{table*}

\begin{table*}[tb]
\centering
\begin{tabular}{|l|c|c|c|c|c|c|}
\hline
\multirow{2}{*}{Model} & \multicolumn{3}{c|}{Test set 1} & \multicolumn{3}{c|}{Test set 2} \\ \cline{2-7}
& P & R & F & P & R & F \\ \hline
Basic seq2seq & 26.03 & 13.52 & 17.80 & 30.97 & 23.70 & 26.85 \\ 
+ coverage & 26.69 & 12.88 & 17.37 & 37.09 & 24.12 & 29.23 \\ 
+ soft loss & 29.3 & 17.26 & 21.72 & 37.90 & 27.63 & 31.96 \\
+ coverage \& soft loss & 29.57 & 17.30 & 21.83 & 38.22 & 30.18 & 33.73 \\ \hline
\end{tabular}
\caption{Ablation results of applying coverage mechanism and soft loss function. Test set 1 and test set 2 are the same as Table \ref{tab:micro Experiment results}}\label{tab: micro Ablation experiment results}
\end{table*}

In Table \ref{tab:micro Experiment results}, we show the results of our proposed models as well as the baseline models. One thing that should be noted is that since the data in Test set 2 (extracted from text book) have much better quality than Test set 1, the performance on Test set 2 is much higher than it is on Test set 1, which is consistent with our instinct. 

From the experiment results we can see that the baseline model multi-label has higher micro recall rate 29.72, 40.49 but much lower micro precision 10.83, 13.51. This is because unlike the seq2seq model that dynamically determines the length of the generated sequence, the output length is rigid and can only be determined by thresholds. We take the tokens within the top 20 as the answer for the multi-label model.

As to the basic seq2seq model, although it beats the multi-label model overall, the recall rate drops substantially. This problem is partly caused by the repetition problem, the basic seq2seq model sometimes predicts high frequent tokens instead of more meaningful ones. Apart from this, although the seq2seq based model is better able to model the correlation between target labels, it makes a strong assumption on the order of the target sequence. In the prescription generation task, the order between herb tokens are helpful for generating the sequence. However, since the order between the herbs does not affect the effect of the prescription, we do not consider the order when evaluating the generated sequence. We call the phenomenon that the herbs are under the ``weak order''. The much too strong assumption on order can hurt the performance of the model when the correct tokens are placed in the wrong order.

In Table \ref{tab: micro Ablation experiment results} we show the effect of applying coverage mechanism and soft loss function. 

% \begin{table*}[tb]
% \centering
% \small
% \setlength{\tabcolsep}{3pt}
% \begin{tabular}{|l|c|c|c|}
% \hline
% Model & Original Total Length & Pruned Total Length & Repetition Number \\ \hline
% Basic seq2seq & 859 & 716 & 143 \\
% + coverage (proposed)  & 724 & 609 & 115 \\
% + soft loss (proposed) & 741 & 685 & 56 \\
% Proposal & 782 & 743 & 39 \\ \hline
% \end{tabular}
% \caption{Prescription length of different models on test set 2. }\label{tab: Repetition number}
% \end{table*}

Coverage mechanism gives a sketch on the prescription. The mechanism not only encourages the model to generate novel herbs but also enables the model to generate tokens based on the already predicted ones. This can be proved by the improvement on Test set 2, where both the precision and the recall are improved over the basic seq2seq model.

The most significant improvement comes from applying the soft loss function. The soft loss function can relieve the strong assumption of order made by seq2seq model. Because predicting a correct token in the wrong position is not as harmful as predicting a completely wrong token. This simple modification gives a big improvement on both test sets for all the three evaluation metrics. 

%In Table \ref{tab: Repetition number} we show the total length of generated prescriptions. We choose the test set 2 because it enjoys better quality. From the table, we can also observe that applying coverage mechanism and soft loss function can greatly reduce the repetition number compared with the basic seq2seq model. 

\subsection{Case Study\label{sec:case_study}}
\begin{table}[t]
\begin{center}
\small
\begin{tabular}{|l|p{5cm}|}
\hline
Symptoms & \begin{CJK*}{UTF8}{gbsn}外感风寒表虚证。恶风发热，汗出头疼，鼻鸣咽干，苔白不渴，脉浮缓或浮弱。\end{CJK*}\\ \hline
Translation & Exogenous wind-cold exterior deficiency syndrome. Aversion to wind, fever, sweating, headache, nasal obstruction, dry throat, white tongue coating, not thirsty, floating slow pulse or floating weak pulse. \\ \hline
Reference & \begin{CJK*}{UTF8}{gbsn}桂枝\ 芍药\ 甘草\ 生姜\ 大枣\end{CJK*}\\
\hline 
Multi-label & \begin{CJK*}{UTF8}{gbsn}防风\ 知母\ 当归\ 川芎\ 黄芪\ 橘红\ 甘草\ 茯苓\ 白术\ 葛根\ 荆芥\ 柴胡\ 麦冬\ 泽泻\ 车前子\ 石斛\ 木通\ 赤茯苓\ 升麻\ 白芍药 \end{CJK*} \\ \hline
Basic seq2seq & \begin{CJK*}{UTF8}{gbsn}柴胡\ 干葛\ 川芎\ 桔梗\ 甘草\ 陈皮\ 半夏 \end{CJK*} \\ \hline
Proposal & \begin{CJK*}{UTF8}{gbsn}桂枝\ 麻黄\ 甘草\ 生姜\ 大枣 \end{CJK*}  \\ \hline
\end{tabular}
\caption{Actual predictions made by various models in test set 2. Multi-label model generates too many herb tokens, so we do not list all of them here. Reference is the standard answer prescription given by the text book.\protect\footnotemark
\label{tab:case study}}
\end{center}
\end{table}

In this subsection, we show an example generated by various models in Table \ref{tab:case study} in test set 2 because the quality of test set 2 is much more satisfactory. The multi-label model produces too many herbs that lower the precision, we do not go deep into its results, already we report its results in the table.

For the basic seq2seq model, the result is better than multi-label baseline in this case. \begin{CJK*}{UTF8}{gbsn}``柴胡''\ (radix bupleuri)、``葛根''\ (the root of kudzu vine)\ can be roughly matched with ``恶风发热，汗出头疼'' (Aversion to wind, fever, sweating, headache), ``甘草''\ (Glycyrrhiza)、``陈皮''\  (dried tangerine or orange peel)、``桔梗''\  (Platycodon grandiflorum)\ can be roughly matched with \ ``鼻鸣咽干，苔白不渴''\ (nasal obstruction, dry throat, white tongue coating, not thirsty), ``川芎''\ (Ligusticum wallichii)\ can be used to treat the symptom of ``头疼''\ (headache)\end{CJK*}. In this case, most of the herbs can be matched with certain symptoms in the textual description. However, the problem is that unlike the reference, the composition of herbs lacks the overall design. The symptoms should not be treated independently, as they are connected to other symptoms. For example, the appearance of symptom \begin{CJK*}{UTF8}{gbsn}``头疼''\end{CJK*} (headache) must be treated together with \begin{CJK*}{UTF8}{gbsn}``汗出''\end{CJK*} (sweat). When there is simply headache without sweat,  \begin{CJK*}{UTF8}{gbsn}``川芎''\end{CJK*} (Ligusticum wallichii) may be suitable. However, since there is already sweat, this herb is not suitable in this situation. This drawback results from the fact that this model heavily relies on the attention mechanism that tries to match the current hidden state in the decoder to a part of the context in the encoder every time it predicts a token.

\footnotetext{Translation: 
\begin{CJK*}{UTF8}{gbsn}桂枝\ - cassia twig, 芍药\ - Chinese herbaceous peony 大黄\ - Rhubarb, 厚朴\ - Magnolia officinalis, 枳实\ - Fructus Aurantii Immaturus, 芒硝\ - Mirabilite, 栀子\ - Cape Jasmine Fruit, 枳壳\ - Fructus Aurantii, 当归\ - Angelica Sinensis, 甘草\ - Glycyrrhiza, 黄芩\ - Scutellaria, 生姜\ - ginger, 大枣\ - Chinese date, 柴胡\ - radix bupleuri, 葛根\ - the root of kudzu vine, 陈皮\ - dried tangerine or orange peel, 桔梗\ - Platycodon grandiflorum, 川芎\ - Ligusticum wallichii, 麻黄\ - Chinese ephedra \end{CJK*}}

For our proposed model, the results are much more satisfactory. \begin{CJK*}{UTF8}{gbsn}``外感风寒''\end{CJK*} (Exogenous wind-cold exterior deficiency syndrome) is the reason of the disease, the symptoms \begin{CJK*}{UTF8}{gbsn}``恶风发热，汗出头疼，鼻鸣咽干，苔白不渴，脉浮缓或浮弱''\end{CJK*} (Aversion to wind, fever, sweating, headache, nasal obstruction, dry throat, white tongue coating, not thirsty, floating slow pulse or floating weak pulse) are the corresponding results. The prescription generated by our proposed model can also be used to cure \begin{CJK*}{UTF8}{gbsn}``外感风寒''\end{CJK*} (Exogenous wind-cold exterior deficiency syndrome), in fact \begin{CJK*}{UTF8}{gbsn}``麻黄''\ (Chinese ephedra)\ and ``桂枝''\ (cassia twig)\end{CJK*} together is a common combination to cure cold. However, \begin{CJK*}{UTF8}{gbsn}``麻黄''\end{CJK*} (Chinese ephedra) is not suitable here because there is already sweat. One of the most common effect of \begin{CJK*}{UTF8}{gbsn}``麻黄''\end{CJK*} (Chinese ephedra) is to make the patient sweat. Since there is already sweat, it should not be used. Compared with the basic seq2seq model, our proposed model have a sense of overall disease, rather than merely discretely focusing on individual symptoms.

From the above analysis, we can see that compared with the basic seq2seq model, our proposed soft seq2seq model is aware more of the connections between symptoms, and has a better overall view on the disease. This advantage is correspondent to the principle of prescribing in TCM that the prescription should be focusing on the \begin{CJK*}{UTF8}{gbsn}``辩证''\end{CJK*} (the reason behind the symptoms) rather than the superficial \begin{CJK*}{UTF8}{gbsn}``症''\end{CJK*} (symptoms).

\section{Conclusion}
In this paper, we propose a TCM prescription generation task that automatically predicts the herbs in a prescription based on the textual symptom descriptions. To our knowledge, this is the first time that this critical and practicable task has been considered. To advance the research in this task, we construct a dataset of 82,044 symptom-prescription pairs based on the TCM Prescription Knowledge Base.

Besides the automatic evaluation, we also invite professionals to evaluate the prescriptions given by various models, the results of which show that our model reaches the score of 7.3 out of 10, demonstrating the effectiveness. In the experiments, we observe that directly applying seq2seq model would lead to the repetition problem that lowers the recall rate and the strong assumption of the order between herb tokens can hurt the performance. We propose to apply the coverage mechanism and the soft loss function to solve this problem. From the experimental results, we can see that this approach alleviates the repetition problem and results in an improved recall rate. %We hope this work can lay a foundation and encourage more researchers to pay attention to the automatic TCM prescription generation problem.

\bibliography{emnlp2018}

\begin{thebibliography}{19}
\expandafter\ifx\csname natexlab\endcsname\relax\def\natexlab#1{#1}\fi

\bibitem[{Bahdanau et~al.(2014)Bahdanau, Cho, and Bengio}]{bahdanau2014neural}
Dzmitry Bahdanau, Kyunghyun Cho, and Yoshua Bengio. 2014.
\newblock Neural machine translation by jointly learning to align and
  translate.
\newblock \emph{arXiv preprint arXiv:1409.0473}.

\bibitem[{Cho et~al.(2014)Cho, van Merrienboer, Gulcehre, Bahdanau, Bougares,
  Schwenk, and Bengio}]{cho2014learning}
Kyunghyun Cho, Bart van Merrienboer, Caglar Gulcehre, Dzmitry Bahdanau, Fethi
  Bougares, Holger Schwenk, and Yoshua Bengio. 2014.
\newblock Learning phrase representations using rnn encoder--decoder for
  statistical machine translation.
\newblock In \emph{Proceedings of the 2014 Conference on Empirical Methods in
  Natural Language Processing (EMNLP)}, pages 1724--1734. Association for
  Computational Linguistics.

\bibitem[{Kingma and Ba(2015)}]{Kingma2015}
Diederik~P. Kingma and Jimmy~Lei Ba. 2015.
\newblock {Adam: a Method for Stochastic Optimization}.
\newblock \emph{International Conference on Learning Representations 2015},
  pages 1--15.

\bibitem[{Li et~al.(2016)Li, Monroe, Ritter, Galley, Gao, and
  Jurafsky}]{li2016deep}
Jiwei Li, Will Monroe, Alan Ritter, Michel Galley, Jianfeng Gao, and Dan
  Jurafsky. 2016.
\newblock Deep reinforcement learning for dialogue generation.
\newblock \emph{arXiv preprint arXiv:1606.01541}.

\bibitem[{Li et~al.(2017)Li, Monroe, Shi, Ritter, and
  Jurafsky}]{li2017adversarial}
Jiwei Li, Will Monroe, Tianlin Shi, Alan Ritter, and Dan Jurafsky. 2017.
\newblock Adversarial learning for neural dialogue generation.
\newblock \emph{arXiv preprint arXiv:1701.06547}.

\bibitem[{Li and Yang(2017)}]{li2017distributed}
Wei Li and Zheng Yang. 2017.
\newblock Distributed representation for traditional chinese medicine herb via
  deep learning models.
\newblock \emph{arXiv preprint arXiv:1711.01701}.

\bibitem[{Mi et~al.(2016)Mi, Sankaran, Wang, and Ittycheriah}]{Mi2016}
Haitao Mi, Baskaran Sankaran, Zhiguo Wang, and Abe Ittycheriah. 2016.
\newblock {Coverage Embedding Models for Neural Machine Translation}.
\newblock \emph{Proceedings of the 2016 Conference on Empirical Methods in
  Natural Language Processing (EMNLP-16)}, (Section 5):955--960.

\bibitem[{Nam et~al.(2017)Nam, Menc{\'\i}a, Kim, and
  F{\"u}rnkranz}]{nam2017maximizing}
Jinseok Nam, Eneldo~Loza Menc{\'\i}a, Hyunwoo~J Kim, and Johannes
  F{\"u}rnkranz. 2017.
\newblock Maximizing subset accuracy with recurrent neural networks in
  multi-label classification.
\newblock In \emph{Advances in Neural Information Processing Systems}, pages
  5419--5429.

\bibitem[{See et~al.(2017)See, Liu, and Manning}]{see2017get}
Abigail See, Peter~J Liu, and Christopher~D Manning. 2017.
\newblock Get to the point: Summarization with pointer-generator networks.
\newblock \emph{arXiv preprint arXiv:1704.04368}.

\bibitem[{Sutskever et~al.(2014)Sutskever, Vinyals, and
  Le}]{sutskever2014sequence}
Ilya Sutskever, Oriol Vinyals, and Quoc~V Le. 2014.
\newblock Sequence to sequence learning with neural networks.
\newblock In \emph{Advances in neural information processing systems}, pages
  3104--3112.

\bibitem[{Tu et~al.(2016{\natexlab{a}})Tu, Lu, Liu, Liu, and Li}]{Tu2016}
Zhaopeng Tu, Zhengdong Lu, Yang Liu, Xiaohua Liu, and Hang Li.
  2016{\natexlab{a}}.
\newblock {Coverage-based Neural Machine Translation}.
\newblock \emph{Arxiv}, pages 1--19.

\bibitem[{Tu et~al.(2016{\natexlab{b}})Tu, Lu, Liu, Liu, and
  Li}]{tu2016modeling}
Zhaopeng Tu, Zhengdong Lu, Yang Liu, Xiaohua Liu, and Hang Li.
  2016{\natexlab{b}}.
\newblock Modeling coverage for neural machine translation.
\newblock \emph{arXiv preprint arXiv:1601.04811}.

\bibitem[{Vinyals et~al.(2015)Vinyals, Bengio, and Kudlur}]{vinyals2015order}
Oriol Vinyals, Samy Bengio, and Manjunath Kudlur. 2015.
\newblock Order matters: Sequence to sequence for sets.
\newblock \emph{arXiv preprint arXiv:1511.06391}.

\bibitem[{Wang(2013)}]{Wang2013TCM}
Liwen Wang. 2013.
\newblock \emph{TCM inquiry modelling research based on Deep Learning and
  Conditional Random Field multi-lable learning methods}.
\newblock Ph.D. thesis, East China University of Science and Technology.

\bibitem[{Wang et~al.(2004)Wang, Qu, Liu, and Cheng}]{wang2004self}
Xuewei Wang, Haibin Qu, Ping Liu, and Yiyu Cheng. 2004.
\newblock A self-learning expert system for diagnosis in traditional chinese
  medicine.
\newblock \emph{Expert systems with applications}, 26(4):557--566.

\bibitem[{Yin et~al.(2015)Yin, Jiang, Lu, Shang, Li, and Li}]{yin2015neural}
Jun Yin, Xin Jiang, Zhengdong Lu, Lifeng Shang, Hang Li, and Xiaoming Li. 2015.
\newblock Neural generative question answering.
\newblock \emph{arXiv preprint arXiv:1512.01337}.

\bibitem[{Zhang(2011)}]{Zhang2011Topic}
Xiaoping Zhang. 2011.
\newblock \emph{Topic Modelling and its application in TCM clinical diagonosis
  and treatment}.
\newblock Ph.D. thesis, Beijing Transportation University.

\bibitem[{Zhipeng et~al.(2017)Zhipeng, Jianqiang, Yingfeng, Fang, and
  Luo}]{Zhu2017TCM}
Zhu Zhipeng, Du~Jianqiang, Liu Yingfeng, Yu~Fang, and Jigen Luo. 2017.
\newblock Tcm prescription similartiy computation based on lda topic modelling.
\newblock \emph{Application Research Of Computers}, pages 1668--1670.

\bibitem[{Zhou et~al.(2010)Zhou, Chen, Liu, Zhang, Wang, Li, Guo, Zhang, Gao,
  and Yan}]{zhou2010development}
Xuezhong Zhou, Shibo Chen, Baoyan Liu, Runsun Zhang, Yinghui Wang, Ping Li,
  Yufeng Guo, Hua Zhang, Zhuye Gao, and Xiufeng Yan. 2010.
\newblock Development of traditional chinese medicine clinical data warehouse
  for medical knowledge discovery and decision support.
\newblock \emph{Artificial Intelligence in medicine}, 48(2):139--152.

\end{thebibliography}
\bibliographystyle{acl_natbib_nourl}

\end{document}